\documentclass[letterpaper,twocolumn,fleqn]{article} 

\usepackage{ist}
\usepackage{algorithm2e}

\title{Resampling Forgery Detection Using Deep Learning and A-Contrario Analysis}

\author{ A. Flenner, L. Peterson; NAVAIR; China Lake, CA\\J. Bunk, T. M. Mohammed, L. Nataraj, B.S. Manjunath; Mayachitra Inc.; Santa Barbara, CA}

\date{} 

\begin{document} 

\maketitle 

\thispagestyle{empty} 


\begin{abstract}
The amount of digital imagery recorded has recently grown exponentially, and with the advancement of software, such as Photoshop or Gimp, it has become easier to manipulate images.  However, most images on the internet have not been manipulated and any automated manipulation detection algorithm must carefully control the false alarm rate.  In this paper we discuss a method to automatically detect local resampling using deep learning while controlling the false alarm rate using a-contrario analysis.  The automated procedure consists of three primary steps.  First, resampling features are calculated for image blocks.  A deep learning classifier is then used to generate a heatmap that indicates if the image block has been resampled.  We expect some of these blocks to be falsely identified as resampled.  We use a-contrario hypothesis testing to both identify if the patterns of the manipulated blocks indicate if the image has been tampered with and to localize the manipulation.  We demonstrate that this strategy is effective in indicating if an image has been manipulated and localizing the manipulations.  
\end{abstract}

\section{Introduction}
\label{sec:intro}

Image tampering techniques have become sophisticated and in many instances experts have difficulty identifying if an image has been modified. Furthermore, with the advent of new cameras, smartphones and tablets, the amount of digital images has grown exponentially and the tools for digitally manipulating these images, such as Photoshop, Gimp, Snapseed, and Pixlr, have evolved significantly making it easy to modify many images in a short time span. Due to these advancements, the field of digital image forensics needs to develop tools that can quickly verify image authenticity and localize the regions where an image has been manipulated. In this paper, we combine our previous works on resampling detection \cite{bunk2017detection} and \textit{a-contrario} analysis \cite{flenner2008two, flenner2011helmholtz} to assign a tamper score and localize the image tampering.  We demonstrate that our algorithm is effective at detecting many different types of image tampering. 

Many categories of image forgeries, including copy-move, splicing, and object removal, often implement resampling as part of the forgery workflow.  Resampling is required in order to blend the inserted or modified regions of the image with the rest of the image.  For this reason, resampling detection is capable of detecting many different types of image manipulations and many resampling detector have been recently proposed~\cite{popescu-farid-resampling, prasad2006resampling, second-diff-method, babak-radon, kirchner-local, feng2012normalized, ryu2014estimation}.  However, all of these resampling detection techniques are not completely accurate for certain image regions and false alarms occur when anti-detection methods are implemented.  There is a need to carefully control the false alarm rate at the image level.      

In \cite{bunk2017detection} we designed a detector to improve on previous resampling detectors.  Our detector found image artifacts imposed by classic upsampled, downsampled, clockwise rotation, counterclockwise rotation, and shearing methods.  We combined these five different resampling detectors with a JPEG compression detector and for each of the six detectors we output a heatmap corresponding to a confidence score of where the image was manipulated.  We observed in our previous work that our heatmaps were noisy, and in \cite{bunk2017detection} we smoothed the heatmaps to localize the detection and determine the detection score. In this work, motivated by \cite{flenner2008two, flenner2011helmholtz}, we develop a new detection scheme that combines two ideas:\textit{a-contrario} statistical hypothesis testing and image segmentation. 

The \textit{a- contrario} methodology was originally formulated by Desolneux, Moison, and Morel in order to rigorously implement the Helmholt Theroy of perception\cite{Desolneux2000}.  Motivated by the lack of principled methods to determine an appropriate decision threshold for many basic computer vision tasks they formulated a structured statistical test to detect image primitives such as line segments and contrasted boundaries.   

We were inspired to use the \textit{a-contrario} procedure since it only requires a model, the background model, for unstructured data, i.e. heatmaps in which no resampling has occurred, and we do not need to postulate a model for heatmaps that indicate resampoing has occurred.  Our procedure starts by scanning the heatmap for all possible detection events, where a detection event is a location in the heatmap plus a group of pixels within the heatmap that makes it a valid detection. Since the background model is the model for unstructured data, then any detection within the event set is a false detection according to the \textit{a-contrario} model and the goal of \textit{a-contrario} modeling is to provide an upper bound on the number of these false detections. 

\section{Previous Work}
The image forensics field recognizes many different categories of image forgeries with copy-move, splicing, and  object removal as some of the most popular.  In many of these forgeries, resampling is a necessary workflow element.  Due to the diversity of image tampering methods, there are numerous techniques to identify if an image has been manipulated and we briefly review some of the techniques below with a special emphasis on those that overlap with our method.  We then review \textit{a-contrario} analysis.  

\subsection{Resampling and Image Forgery Detection}
Resampling an image requires an interpolation method and linear or cubic interpolations are very popular and this fact was exploited by the authors of~\cite{popescu-farid-resampling}.  They implemented an Expectation-Maximization (EM) algorithm to detect  periodic correlations introduced by interpolation.   However the periodic JPEG blocking artifacts also introduce periodic patterns that confuses their resampling detector.
The variance of the second difference operator was used by ~\cite{second-diff-method} to detect resampling on images that are scaled using linear or cubic interpolations. Their method is most efficient at detecting up-scaling and it is very robust to JPEG compression with detection possible even at very low quality factors (QF). Downscaled images can be detected but not as robustly as upscaled images.  In ~\cite{babak-radon}, the Radon transform and a derivative filter was exploited to improved the quality of the results and to address other forms of resampling.  In~\cite{ kirchner-local}, a simpler method than~\cite{popescu-farid-resampling} was derived by using a linear predictor residue instead of the computationally expensive EM algorithm.  This resampling body of work motivated us in~\cite{bunk2017detection} where we combined the linear predictor strategy with deep learning based models in order to detect tampered image patches.

Other resampling detectors include~\cite{ryu2014estimation, Nataraj10-345,Nataraj09-331, feng2011energy,feng2012normalized, golestaneh2014algorithm, kwon2015efficient}.  In~\cite{ryu2014estimation} periodic properties of interpolation were found using the second-derivative of the image and these properties were used for detecting image manipulation.  Resampling on JPEG compressed images was detected in~\cite{Nataraj10-345,Nataraj09-331} by added noise before passing the image through the resampling detector and they showed that adding noise improved resampling detection. 
A normalized energy feature was implemented in~\cite{feng2011energy,feng2012normalized} and a support vector machine (SVM) was subsequently used to classify resampled images. 
Furthermore, recent approaches to reduce the effects of JPEG artifacts were developed in~\cite{golestaneh2014algorithm, kwon2015efficient}. 

We tested our resampling detector on a broad range of general image tampering techniques.  Most image tamper detection strategies are directed toward a specific type of image forgery such as copy-move or object insertions and removals and we review some common approaches.  For copy-move forgeries a common approach is to match image features within the image.  In order to detect copy-move forgeries, an image is first divided into overlapping blocks and some sort of distance measure or correlation is used to determine blocks that have been cloned. For example, in~\cite{amerini2011sift} copy-move forgeries were detected using SIFT features.  Many similar methods to detect copy-move have been proposed~\cite{li2015segmentation,kakar2012exposing, jaberi2014accurate,al2013passive}.

Another strategy to detect copy move forgeries is to match a transformation of image regions rather than image regions themselves.  In \cite{fridrich2003detection}, Fridrich \textit{et. al} use DCT coefficients of image regions to help find duplicate DCT blocks while Popsecu and Farid used PCA \cite{popescu-farid-tr-04} to detect duplicated regions and Mahdian and Saic use a combination of blur invariant moments and PCA \cite{mahdian2007detection}.  A matching image regions to detect copy-move forgeries becomes more difficult if the moved region undergoes some transformation such as scalings that makes region matching difficult.  Bayram et. al \cite{bayram2009efficient} addresses this issue by using a combination of Fourier Mellin transforms, which are invariant to rotation, scale and translation, and Bloom filters. 
Another issue in locating copy-move forgeries is the computational time to find matching patches.  
In \cite{cozzolino2015efficient},  the patch-match algorithm is used to efficiently compute an approximate nearest neighbor field over an image.
They added robustness to their algorithm by using invariant features such as Circular Harmonic transforms and show that they can detect duplicated blocks that have undergone geometrical transformations and then perform keypoint matching.  
  
In~\cite{manu2015visual}, an image splicing detection technique has been proposed using visual artifacts. 
A novel image forgery detection method is presented in~\cite{muhammad2014image} based on the steerable pyramid transform (SPT) and local binary pattern (LBP). 
The paper~\cite{guillemot2014image} includes the recent advances in image manipulation and discusses the process of restoring missing or damaged areas in an image. 
In~\cite{ansari2014pixel}, the authors review the different image forgery detection techniques in image forensic literature. 

In computer vision, deep learning shows outstanding performance in different visual recognition tasks such as image classification~\cite{zhou2014learning}, and semantic segmentation~\cite{long2015fully}.  For this reason, there has been a growing interest to detect image manipulation by applying different computer vision and deep-learning algorithms~\cite{bayar2016deep, bayar2017design,rao2016deep,bappy2017exploiting}.  In~\cite{long2015fully}, two fully convolution layers have been exploited to segment different objects in an image. The segmentation task has been further improved in \cite{zheng2015conditional, badrinarayanan2017segnet}. These models extract hierarchical features to represent the visual concept, which is useful in object segmentation. Since, the manipulation does not exhibit any visual change with respect to genuine images, these models often do not perform well in segmenting manipulated regions. 

Other deep learning methods include detection of generic manipulations~\cite{bayar2016deep, bayar2017design}, resampling~\cite{bayar2017resampling}, splicing~\cite{rao2016deep} and bootleg~\cite{buccoli2014unsupervised}.   In~\cite{qian2015deep} the authors propose Gaussian-Neuron CNN (GNCNN) for steganalysis detection.
A deep learning approach to identify facial retouching was proposed in~\cite{bharati2016detecting}.
In~\cite{zhang2016image}, image region forgery detection has been performed using a stacked auto-encoder model. 
In\cite{bayar2016deep},  a new constrained convolutional layer is proposed to learn the manipulated features  from an image. In~\cite{bunk2017detection} an unique network exploiting convolution layers along with LSTM network was presented.

\subsection{A-Contrario Analysis}
Our approach to determining an image manipulation score and localizing manipulated regions is motivated by  \textit{a-contrario} change detection algorithms.  Many change detection algorithms propose to find differences between two images by registering the two images and then finding the magnitude of the difference between the two images.  Registration error plus illumination differences introduce mistakes in such change detection algorithms, and thus controlling false alarms are a priority.  In this work, we replace the difference between the two images with a heatmap generated using our resampling detector.  Since our procedure is a modified change detection algorithm, we review the different \textit{a-contrario} change detection methods.  Furthermore, for simplicity of discussion and to tie in with this work, we refer to the image difference in the change detection algorithms as a heatmap.  

\textit{A-contrario} methods were popularized in image processing by Desolneux, Moisan and Morel, and this procedure was initially applied to identifying edges, alignments, one dimensional histogram modes, and highly contrasted boundaries  \cite{Desolneux2007,Desolneux22001,Desolneux22003,Desolneux32003,Desolneux12004,Desolneux2000}.  Many problems have been addressed using the \textit{a-contrario} methods other than change detection.  These applications include, but are not limited to, automated color segmentation \cite{Delon12007}, mode detection of two dimensional histograms \cite{flenner2008two}, change detection \cite{flenner2011helmholtz}, clustering \cite{muse2006}, and matching local features \cite{Rabin2009}. Fundamental to \textit{a-contrario} methods is the identification of a background probability model $\mathcal{H}_0$, a set of detection events $\mathcal{E}$, and a multiple hypothesis testing procedure that bounds the number of false alarms.    

The earliest work on \textit{a-contrario} methods for change detection is by Lisani and Morel \cite{Lisani2003} where they use the binomial distribution as the background model and a sliding window to determine events.  Their method is sensitive to the the window size and cannot locate changes of arbitrary shapes. The adapted windows technique of Dibos \textit{et al.} \cite{Dibos2009} uses \textit{a-contrario} methods to detect changes in moving video.  They are interested in real time processing, and therefore they use a sliding window method. The use of sliding windows reduces processing time, but it also has less precise boundary information.  Veit \textit{et al.} \cite{Veit12006} used the \textit{a-contrario} methods with the binomial distribution as the background model and level sets with highly contrasted boundaries to determine possible events.  Using contrasted boundaries allows one to find arbitrary shapes in the image, but the number of events is limited. The \textit{a-contrario} method in \cite{flenner2011helmholtz} is built on these ideas but used every possible image level set to determine the set of events.  The paper by Buades \textit{et al.} also explores using the \textit{a-contrario} methods for change detection applications \cite{Buades2009}.  They also use sliding windows and histogram analysis is used to determine the heatmap.  
\begin{figure}[!hb]
	\includegraphics[width=\columnwidth]{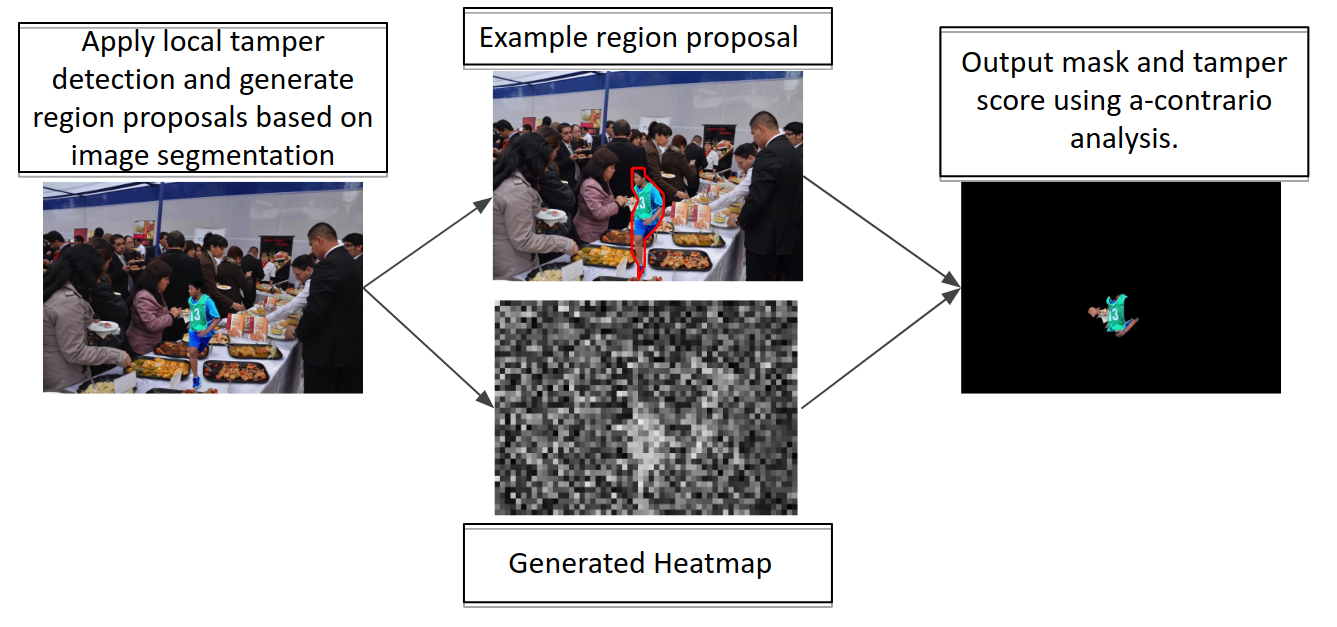}
	\caption{The algorithm workflow. First, a heatmap that indicates resampling is generated using the deep learning models in Figure 2.  A-contrario analysis, illustrated in the last two images, is then used to localize the resampling and determine a total image score. }
	\label{Figure:workflow}
\end{figure}

\section{Methods and Results}
Our procedure to determine manipulated regions consists of two primary steps summarized in Fig. \ref{Figure:workflow}.  First, a neural net classifier is applied to resampling features to determined six different heatmaps.  The heatmap generation was first reported in \cite{bunk2017detection} and we leave the details of the computations to that paper.  Second, \textit{a-contrario} analysis is applied to each heatmap and an image score is obtained using information from all the heatmaps.  The complete \textit{a-contrario} algorithm is given in \cite{flenner2008two} and only an outline of the algorithm will be presented here.   

\subsection{Resampling Heatmap Generation}
\label{sec:heatmap}
\begin{figure}[!ht]
\centering
\includegraphics[width=\columnwidth]{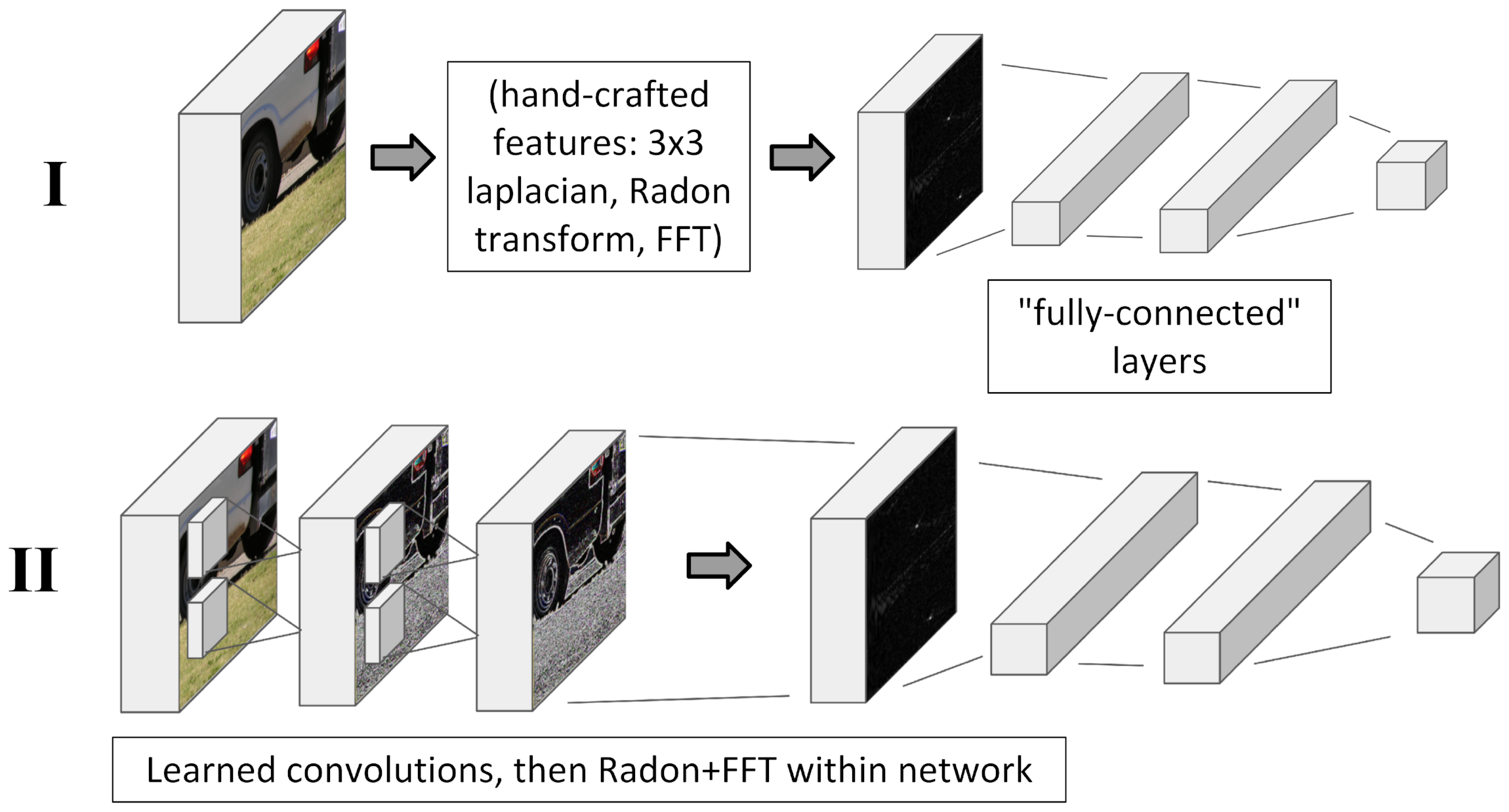}
\vspace{-15pt}
\caption{Deep Neural networks for detecting resampling in small patches.} 
\label{fig:resamp-nets}
\vspace{-10pt} 
\end{figure}

Our initial step is to generate heatmaps that indicate the five different resamplings required for upsampling, downsampling, shearing, clockwise rotation and counterclockwise rotation.  Furthermore, we included a JPEG compression classifier that determines if the input has been compress at a JPEG quality factor of 85 or less.  The five resampling and one JPEG classifier generates six different heatmaps.  
  \begin{figure}[!hb]
  	\centering
  	\includegraphics[width=8 cm]{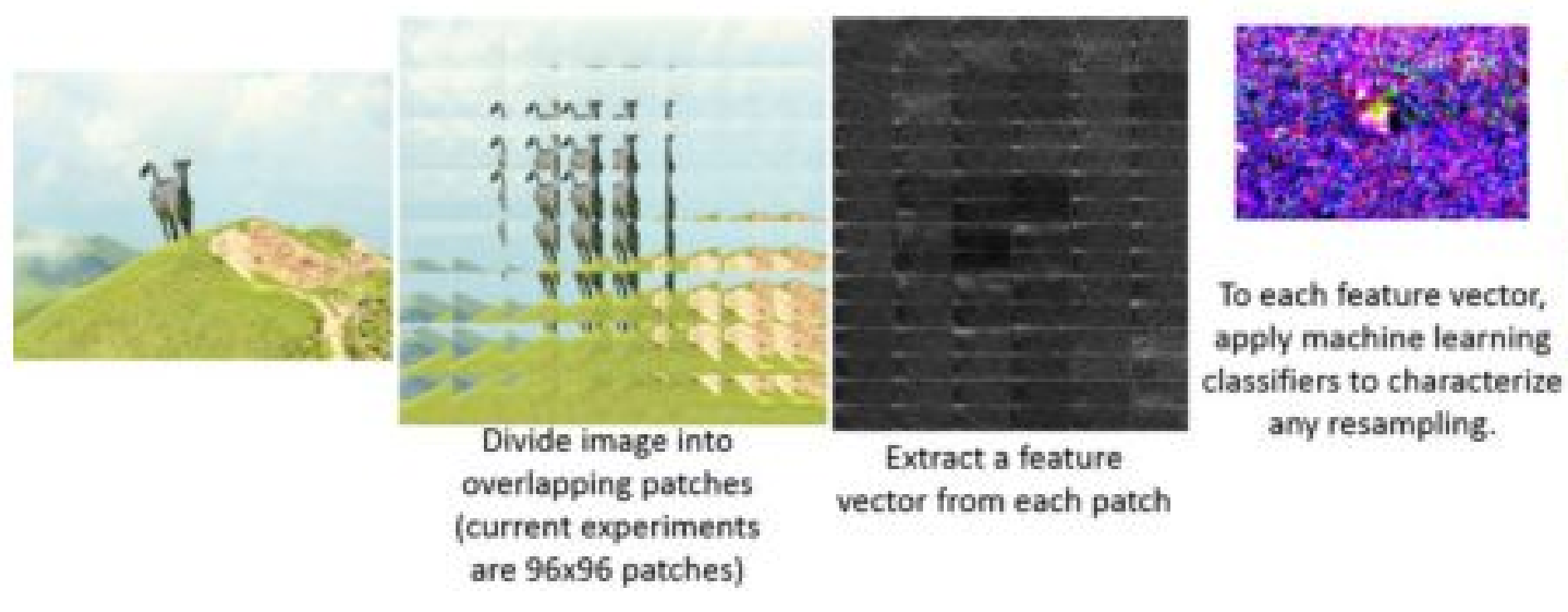}
  	\caption{Patch Feature Extraction based on the Radon Transform} 
  	\label{fig:patchfeat}
  \end{figure}
 In order to train our six heatmaps we use two different procedures.  The first procedure is to extract resampling features from the image and use a deep convolutional neural network to train a classifier on these features.  The second procedure is to train a classifier on the image itself and Fig. \ref{fig:resamp-nets} summarizes the two procedures labeled model I and model II.  In particular, we use model I for rescaling down, rotate clockwise, rotate counterclockwise, and shearing while model II was used for rescaling up and jpeg quality detection.  We chose the model by training both models for all manipulation types and picking the model that gave the best area under the curve (AUC) value on a receiver operating characteristics (ROC) curve.  
 
     In order to localize the manipulations, we do not train the classifiers on the entire image, but rather train the classifiers on 64x64 image patches.  In order to generate features we apply a 3x3 Laplacian filter to these patches and output the linear predictive error magnitude of the patches.  Correlations in the linear predictor error are then found using the Radon transform followed by a 2-D FFT. We trained the classifiers using uncompressed images from the UCID dataset~\cite{schaefer2003ucid} and RAISE data set~\cite{dang2015raise} where we simulated the different resampling methods and JPEG compression. Both data sets were disjoint from our final testing data set.  Note that the classifiers are not mutually exclusive and we did not use a multiclass model but trained each classifier separately.  More details about patch size selection and algorithm design can be found in~\cite{bunk2017detection}.  Note that since the classifier classifies image blocks with a pixel stride of nine the resulting heatmap's width and height is smaller than the original image.  

\subsection{Tamper Score and Mask determination using a-contrario analysis}

    Once the heatmap is generated, \textit{a-contrario} analysis is used to simultaneously determine an image score and to localize the tamper areas.  This procedure requires a background probability model for when no tampering occurs, which we will call $\mathcal{H}_0$, a set of possible manipulation events $\mathcal{E}$, and a decision function based on the background model whose domain are the events.  
    
    Recall that the model $\mathcal{H}_0$ serves as a model when no structure exists within the data, and the set of detection events are possible locations and grouping that correspond to data structures.  For example, in line segment detection, locations are every pixel in the image and structures are straight lines in the image.  Given any event $E \in \mathcal{E}$, we can calculate the probability of $E$ occurring under the hypothesis $\mathcal{H}_0$. Given any probability threshold, we should expect to find false alarms.  The definition of meaningful events bounds the expected number of false alarms by the number one:  

\noindent
\textbf{Definition:} Given a probability model $\mathcal{H}_0$, the events in $\mathcal{E}$ are called meaningful events if the $\mathcal{H}_0$-expected number of events observed is less than or equal to one.

Note that the cardinality of $\mathcal{E}$ can be large, so this technique must correct for multiple comparisons.  Furthermore, due to the large number of event calculations required in a typical applications, it is common to use a large deviation bound in computations and we used the H\"offding bound for Bernoulli random variables in our work \cite{hoeffding1963probability}.  

The \textit{a-contrario} procedure has two inputs.  The first input is one of the six heatmaps where each tested image block is mapped to a tampering confidence score between zero and one where one means high confidence that the image was manipulated.  Let $M$ indicate this mask.  The second input is the original image $I$.  The \textit{a-contrario} procedure thresholds the original mask $M$ to create a binary 0-1 mask and then refines the mask using region proposals derived from the original image.   Let $TM$ be the mask after applying the threshold. Two region proposals were used in the analysis: level set and deep mask proposals. 
    \begin{figure}[!h]
\centering
\includegraphics[width=\columnwidth]{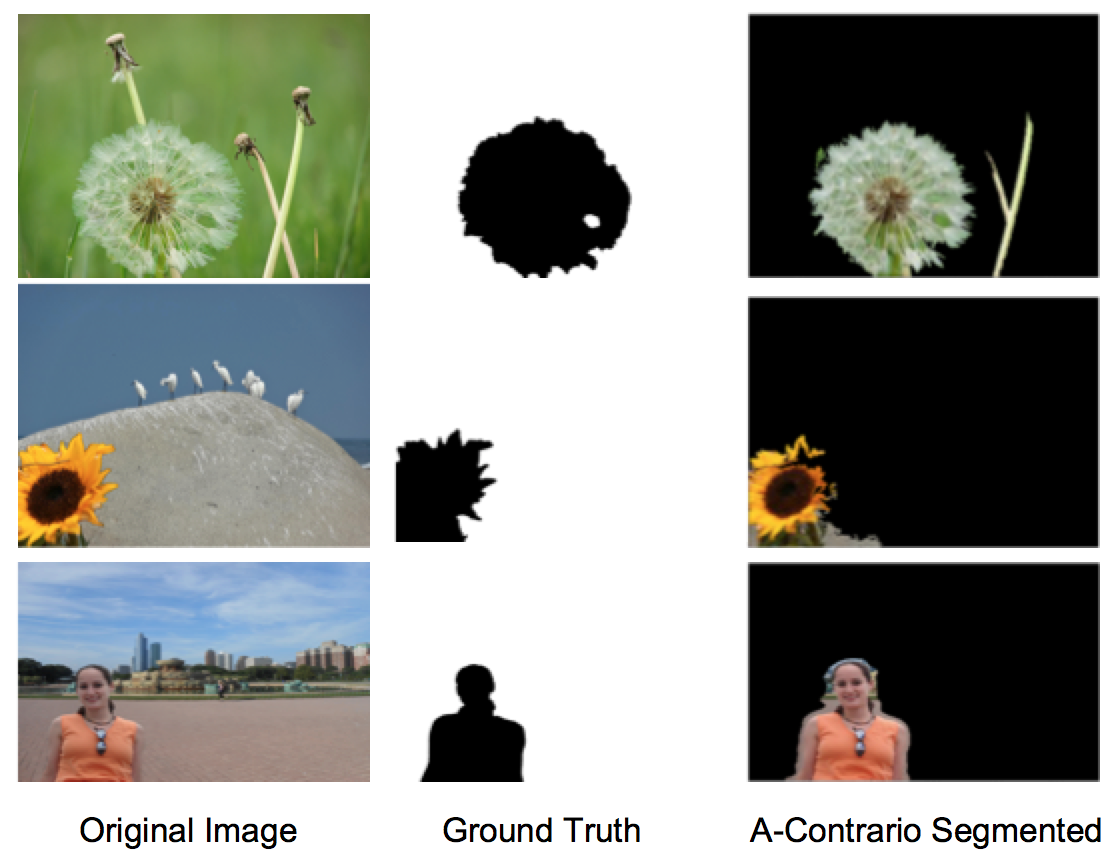}
\vspace{-5pt}
\caption{Example results of the a-contrario segmentation.} 
\label{fig:ac-results}
\vspace{-10pt} 
\end{figure}
    Given the two inputs, the initial step is to threshold the heatmap resulting in a binary mask of zeros and ones with one indicating the block was tampered.  We then assume a Bernouilli model where $p$ is the probability of a heatmap pixel is one. The value of $p$ is determined by counting the number of pixels in the heatmap that are one and dividing by the total number of pixels. In other words, in order to determine if a set is spatially grouped, we assume that the pixels of $TM$ are uniformly distributed on the image and look for regions that do not agree with that hypothesis.  This forms our background model $\mathcal{H}_0$
    
    Given this binary mask, we assume tampering occurs when pixels in $TM$ cluster spatially.  Ideally, we would search every possible connected group of pixels to find clusters of manipulated patches, but this is computationally in-feasible.  We therefore use the fast level set transform \cite{monasse2000fast} and deep mask~\cite{pinheiro2015learning} to segment the image image.  The output of deepmask is a set of connected pixels that correspond to a semantically segmented object in the image.  The output of the fast level set transform is a connected region where the boundary of the region has constant pixel value. We use both methods since deepmask is not trained on all possible regions that we may want to segment, therefore just using deepmask results in missed detection, and the fast level set transform will not always find semantically reasonable regions.   
    
    The segmented image regions are used to limit the number of connected regions required to search.  An event in $\mathcal{E}$ is therefore the region determined by the image together with the count of pixels that are one and the count of pixels that are zero inside the these regions.   

\begin{figure}[!hb]
\centering
	\includegraphics[width=\columnwidth]{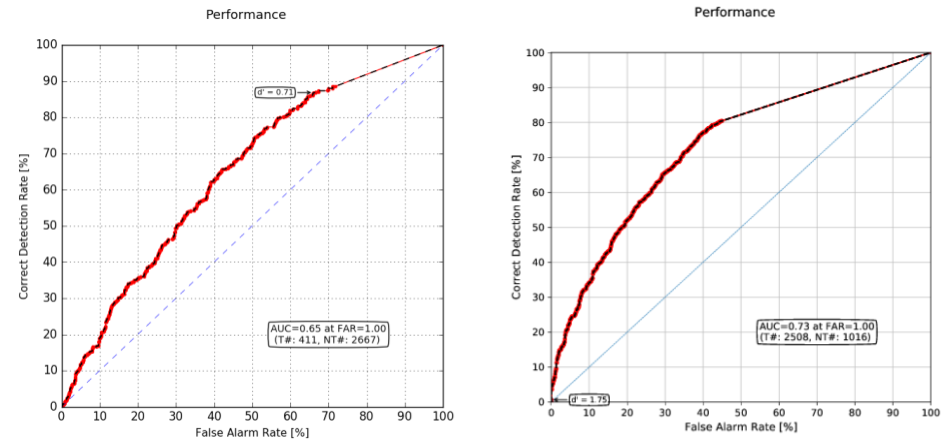}
	\caption{ROC curve for \textit{a-contrario} analysis on the right compared with the method in \cite{bunk2017detection} on the left. }
	\label{Figure:roc}
\end{figure}
\begin{algorithm}[!h]
\KwData{Resampling Heatmap $M$, Original Image $I$}
\KwResult{Segmented Image}
$\{S\} \leftarrow FLST(I) \cup DeepMask(I)$ \\
$TM \leftarrow M > c$ \\
$L \leftarrow $ empty list. \\
\ForAll{$S \in \{S\}$}{
    $r(S) = \sum_{(i,j) \in S} TM(i,j)$ \\
    $NFA(S) = \#S\; \mbox{Pr}(R > r(s))$ \\
    \If{$NFA(S) < 1$}{
        $L.append(S)$
    }
}
Cull $L$ to a list of disjoint sets. \\
Output $L$.
\caption{Pseuodocode to generate a score and mask from each heatmap.}
\end{algorithm}
  Let $S$ be a region proposal, i.e. a segmented region of the image, obtained either through the level set analysis or through the deep mask algorithm. We use a Bernoulli process to determine if the region has been tampered by counting the number of tampered blocks within the region using the formula $r(S) = \sum_{(i,j) \in S} TM(i,j)$.  We then calculate the decision function $NFA(S) = \#S\,\mbox{Pr}(R > r(s))$ where $\#S$ is the total number of region proposals and the probability is the tail of the binomial distribution with probability of success $p$.  Note that $NFA(S)$ stands for the number of false alarms and using the $a-contrario$ model all meaningful events are unlikely to occur due to chance since $NFA(S)$ is an upper bound on the expected number of false alarms.  Each meaningful region satisfies $NFA(S) < 1$ and is flagged as a potentially manipulated region.  The regions are then further refined by determining a disjoint set of regions that have the minimal $NFA(S)$ values. See \cite{flenner2011helmholtz} for a discussion on how to obtain the disjoint regions.

Following this procedure, a score and heatmap is generated for each of the six heatmaps creating a six dimensional vector with each element between zero and one.  A final image score was derived by averaging the non-zero elements of this vector.  If all the elements were zero, the output score was zero.  The final mask is the union of the masks obtained from the non-zero elements.  Example mask results are shown in Fig. \ref{fig:ac-results}.  

We evaluated the proposed model on NIST Nimble 2017 dataset for the Media Forensics challenge. 
This dataset includes mainly three types of manipulation: (a) copy-clone, (b) removal, and (c) splicing. 
The images are tampered in a sophisticated way to beat current state-of-the-art detection techniques.
The results in Fig.~\ref{Figure:roc} show the efficacy of the proposed model for different images in Nimble 2016 dataset. Compared with the results from \cite{bunk2017detection}, where the same feature set was used but a different image scoring method was implemented, the area under the curve metric on the ROC curve increased by .08 using the \textit{a-contrario} method.  

\section{Conclusions and Future Work}
In this paper we described a technique to determine and image tamper score and a mask using resamping features, deep neural networks, and \textit{a-contrario} analysis.  We demonstrated that the \textit{a-contrario} procedure can increase the AUC measurement over previous work.  We note that the \textit{a-contario} procedure is applicable to all heatmaps bounded between zero and one and does not require the resampling heatmaps presented in this paper.

There is room to improve the \textit{a-contrario} procedure presented here through parameter tuning and better integration of the output scores from the heatmaps.  In particular, we applied a threshold to the heatmaps using a value of 0.75 and did not attempt to optimize this threshold. Furthermore, we used a simple mean value to combine the information from the heatmaps.  As mentioned above, the heatmaps are not independent and a more careful statistical analysis should obtain a better final output score.   

\section{Acknowledgments} 

This research was developed with funding from the Defense Advanced Research Projects Agency (DARPA).
The views, opinions and/or findings expressed are those of the author and should not be interpreted as representing the official views or policies of the Department of Defense or the U.S. Government. 
The paper is approved for public release, distribution unlimited.


{\small
	\bibliographystyle{spiejour}
	\bibliography{resamp-acontrario.bbl}
}


\begin{biography}
\textbf{Dr. Arjuna Flenner} received his Ph.D. in Physics at the University of Missouri-Columbia located in Columbia MO in the year 2004. His major emphasis was mathematical Physics. After obtaining his Ph.D., Arjuna Flenner obtained a job as a research physicist for NAVAIR at China Lake CA. He won the 2013 Dr. Delores M. Etter Navy Scientist and Engineer award for his work on Machine Learning. 

\textbf{Lawrence A. Peterson} has extensive research and development experience as a civilian engineer / scientist with the U. S. Navy for over 37 years. He currently heads the  Image and Signal Processing Branch in the Physics and Computational Sciences Division,  Research Directorate at the Naval Air Warfare Center, Weapons Division (NAWCWD), China Lake, CA.  Additional development experience includes signals and imagery exploitation, multi-modal target detection and integrated guidance applications, software development for radar and ESM systems, IR/EO and RF propagation modeling and simulation. 

\textbf{Jason Bunk} received his B.S. degree Computational Physics from the University of California, San Diego in 2015, and the M.S. degree in Electrical and Computer Engineering from the University of California, Santa Barbara in 2016. 
He is currently a Research Staff Member at Mayachitra Inc., Santa Barbara, CA. 
His recent research efforts include applying deep learning techniques to media forensics, and active learning with neural networks for video activity recognition.

\textbf{Tajuddin Manhar Mohammed} received his B.Tech (Hons) degree in Electrical Engineering from Indian Institute of Technology (IIT), Hyderabad, India in 2015 and his M.S. degree in Electrical and Computer Engineering from University of California Santa Barbara (UCSB), Santa Barbara, CA in 2016. After obtaining his Masters degree, Tajuddin Manhar Mohammed obtained a job as a Research Staff Member for Mayachitra Inc., Santa Barbara, CA. His recent research efforts include developing deep learning and computer vision techniques for image forensics and cyber security.

\textbf{Lakshmanan Nataraj} recieved his B.E degree in Electronics and Communications Engineering from Anna university in 2007, and the Ph.D. degree in the Electrical and Computer Engineering from the University of California, Santa Barbara in 2015. 
He is currently a Research Staff Member at Mayachitra Inc., Santa Barbara, CA. 
His research interests include malware analysis and image forensics.

\textbf{B. S. Manjunath} (F’05) received the Ph.D. degree
in electrical engineering from the University of
Southern California, Los Angeles, CA, USA,
in 1991. He is currently a Professor of Electrical and
Computer Engineering with the University of
California at Santa Barbara, Santa Barbara, CA,
USA, where he directs the Center for Multi-Modal
Big Data Science and Healthcare. He has authored or
co-authored about 300 peer-reviewed articles. He is
a Co-Editor of the book entitled Introduction to
MPEG-7 (New York: Wiley, 2002). His current research interests include image processing, computer vision, biomedical image analysis, and large-scale
multimedia databases.

\end{biography}

\end{document}